\documentclass[letterpaper, 10 pt, conference]{ieeeconf}
\usepackage{cite}
\usepackage[font=small]{caption}
\usepackage{amsmath,amssymb,amsfonts}
\usepackage{algorithmic}
\usepackage{graphicx}
\usepackage{textcomp}
\usepackage{xcolor}
\usepackage{subcaption}
\usepackage{hyperref}
\DeclareMathOperator{\diag}{diag}
\newcommand{\loss}{\mathcal{L}}

\DeclareMathOperator{\EX}{\mathbb{E}}

\def\BibTeX{{\rm B\kern-.05em{\sc i\kern-.025em b}\kern-.08em
    T\kern-.1667em\lower.7ex\hbox{E}\kern-.125emX}}

\IEEEoverridecommandlockouts
    
\begin{document}

\title{Modeling Kinematic Uncertainty of Tendon-Driven Continuum Robots via Mixture Density Networks}

\author{Jordan Thompson, Brian Y. Cho, Daniel S. Brown, and Alan Kuntz
\thanks{The authors are with the Robotics Center and the Kahlert School of Computing at the University of Utah, Salt Lake City, UT 84112, USA; email: \{jordan.thompson, brian.cho, daniel.brown, alan.kuntz\}@utah.edu.}
\thanks{This material is based upon work supported in part by the National Science Foundation under grant number 2133027. Any opinions, findings, and conclusions or recommendations expressed in this material are those of the authors and do not necessarily reflect the views of the NSF.}
}

\maketitle

\begin{abstract}
Tendon-driven continuum robot kinematic models are frequently computationally expensive, inaccurate due to unmodeled effects, or both.
In particular, unmodeled effects produce uncertainties that arise during the robot's operation that lead to variability in the resulting geometry.
We propose a novel solution to these issues through the development of a Gaussian mixture kinematic model. We train a mixture density network to output a Gaussian mixture model representation of the robot geometry given the current tendon displacements. This model computes a probability distribution that is more representative of the true distribution of geometries at a given configuration than a model that outputs a single geometry, while also reducing the computation time. We demonstrate one use of this model through a trajectory optimization method that explicitly reasons about the workspace uncertainty to minimize the probability of collision.
\end{abstract}

\section{Introduction}

Much work is being done currently to develop tendon-driven continuum robots for minimally-invasive surgery \cite{wang2021fiora,kato2014tendon,burgner2015continuum}.
However, application of these robots are bottlenecked in part by an inability to accurately and efficiently compute their kinematics.
Stochastic and unmodeled effects on the kinematics of tendon-driven continuum robots may lead to these robots being unsafe to use during surgical tasks as small deviations from the expected geometry could cause unwanted collisions with fragile structures within the body.
Further, as the complexity of the kinematic models grow with the introduction of these effects, so too does the required computation time.
There is a need for a model that is capable of reasoning over these kinematic uncertainties while also minimizing the necessary computation time.
We propose a novel approach to kinematic modeling for tendon-driven surgical manipulators through learned Gaussian mixture models, enabling direct estimation of learned kinematic uncertainty.

Current state-of-the-art physics-based models attempt to explicitly model the different forces acting on the robot \cite{webster2010design, huang20193d, rucker2011statics, grazioso2019geometrically, bentley2022interactive}.
While in theory, these models should accurately compute the geometry of the robot, they tend to either be too restrictive in their assumptions or computationally too slow for real-time surgical applications.
As the complexity of the physics being modeled increases, these models increase in accuracy.
However, they suffer in computation time and in the presence of additional unmodeled or stochastic effects.
Data-driven approaches attempt to solve this problem by implicitly learning how to model the forces that impact the resulting robot geometry \cite{parvaresh2022dynamics, reinhart2016hybrid}.
These approaches decrease the computation time; however, they do not currently possess the ability to handle the inherent uncertainties associated with the robot's kinematics. Any unmodeled effects on the kinematics lead to geometric uncertainties that impact the safe use of these robots. In this paper, we provide results that demonstrate how these unmodeled effects manifest for tendon-driven continuum robots. The issues that arise from the current state-of-the-art models make using them in surgical applications potentially unsafe and/or too slow to perform real-time computations.

We present a solution to modeling the kinematics of tendon-driven continuum robots that can reason implicitly over modeling uncertainty. We train a mixture density network that outputs Gaussian mixture models of the robot's geometry at a given configuration of tendon displacements. Contrary to standard kinematic models that explicitly output the robot's geometry, our method enables reasoning over the probability that the robot will occupy different regions of its workspace. We compare the accuracy of our model across a range of different numbers of mixture components in the model outputs, and we demonstrate a decrease in computation time when compared to a current state-of-the-art Cosserat rod model.

\begin{figure}
    \centering
    \includegraphics[width=\columnwidth]{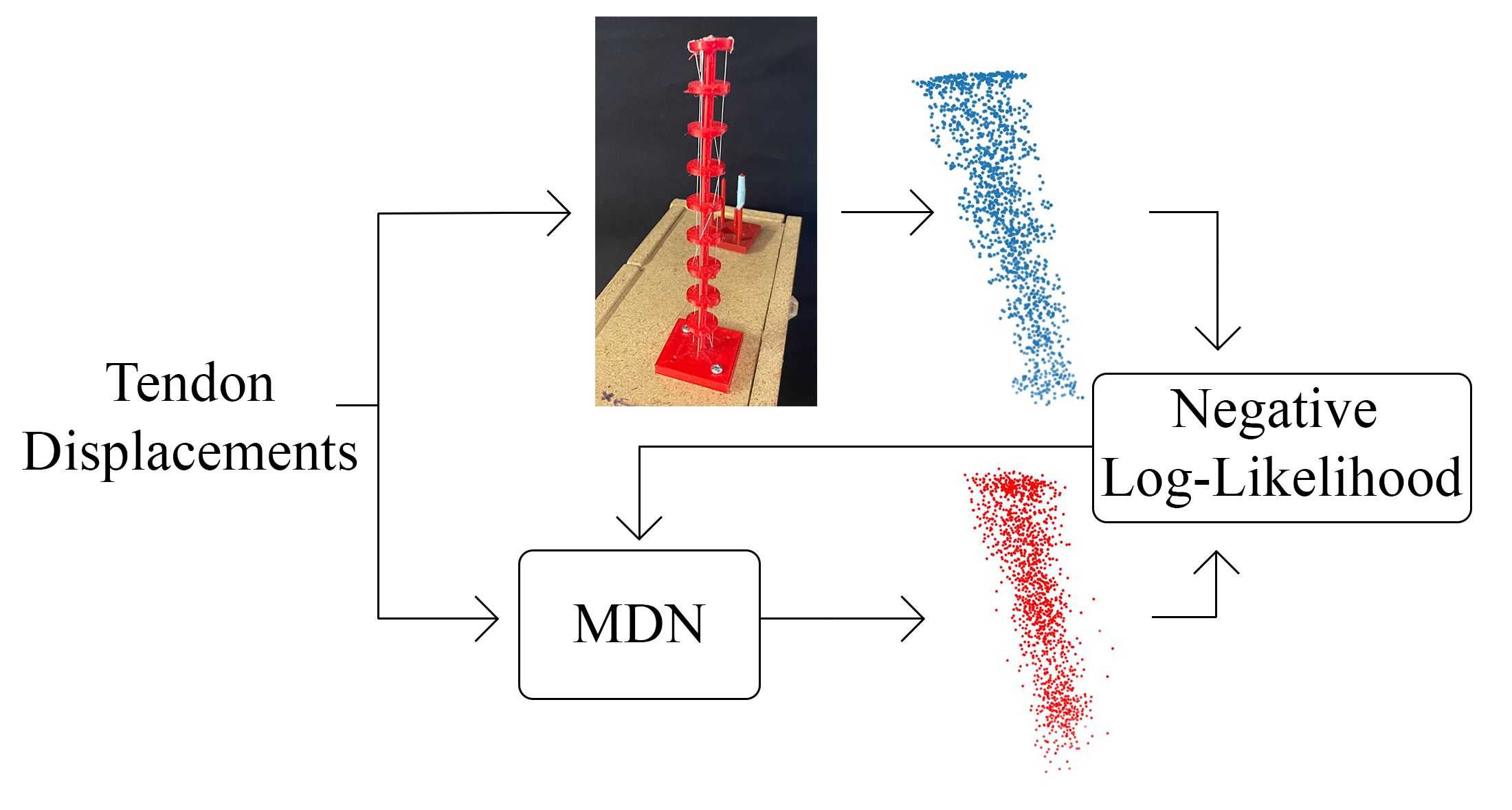}
    \caption{A Mixture Density Network (MDN) is trained to produce a Gaussian mixture model representation of the robot's geometry given the tendon displacements. The negative log-likelihood is computed between the MDN's output and ground-truth point clouds collected from the physical robot. The MDN is then trained to minimize the negative log-likelihood.}
    \label{fig:MainFig}
\end{figure}

We apply our model to an optimization-based motion planner that explicitly minimizes the probability of collision with obstacles along a nominal trajectory. We perform Bayesian optimization with respect to the probability that the robot will collide with the environment. We estimate this probability by integrating the outputs of our model over the obstacles in the environment. We show that using our model enables a reduction in the probability of collision within a simulated chest cavity environment.

\section{Related Work}
\subsection{Physics-Based Models}
Tendon driven continuum robots are subject to many uncertainties, nonlinearities, and disturbances that negatively impact modeling accuracy. Much work has been done to handle these issues through improvements to physics-based kinematic models. Constant Curvature (CC) is one such model that assumes the robot consists of a finite set of curved links where each link can be defined by a set of arc parameters \cite{webster2010design}. This assumption greatly simplifies the kinematics of the robot at the cost of accuracy. Pseudo-rigid body models (PRBM) have also been used to model continuum robot kinematics by assuming the robot consists of a set of rigid links connected by a torsional spring \cite{huang20193d}. PRBM models provide higher accuracy than CC models at the cost of increased computation time.

\begin{figure}
    \centering
    \includegraphics[width=0.8\linewidth]{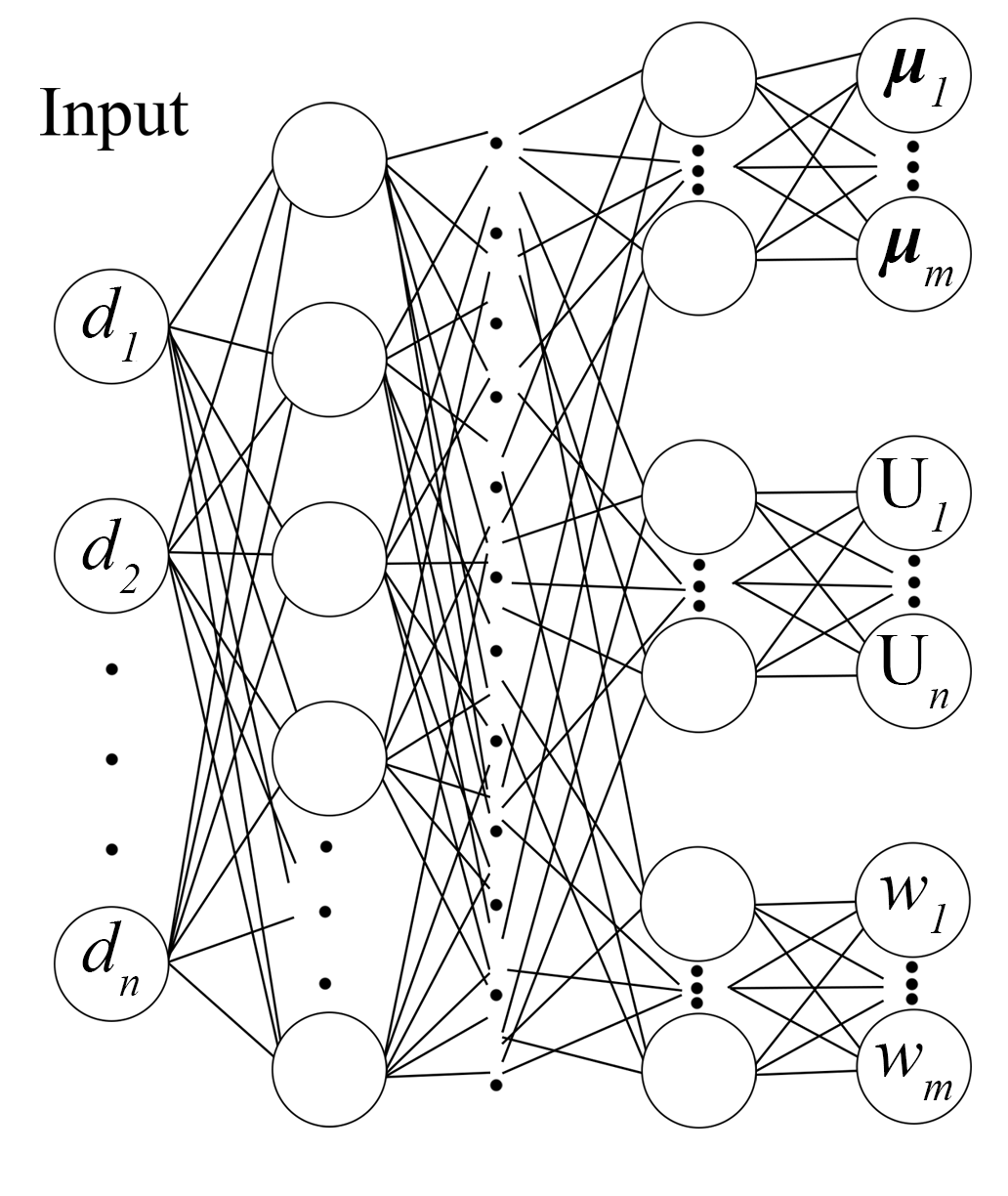}
    \caption{An example mixture density network architecture that takes the list of tendon displacements $d=(d_1,d_2,...,d_n)$ as input and outputs $\boldsymbol{\mu}_i$, $\mathrm{U}_i$, and $w_i$ independently for each component of the resulting Gaussian mixture model.}
    \label{fig:MDNArchitecture}
\end{figure}

As opposed to CC, Variable Curvature (VC) methods such as Cosserat rod models are used to more exactly represent the backbone structure of the robot \cite{rucker2011statics}. The Cosserat rod model assumes the backbone to be a 1-dimensional rod and models the bending, stretching, twisting, and shearing forces applied to the robot. The result is a set of nonlinear ordinary differential equations for the kinematics and a set of nonlinear partial differential equations for the dynamics \cite{grazioso2019geometrically}. While VC models provide a more complete representation of continuum robot geometry than CC models, VC models are generally computationally more expensive. Recent work has provided significant computation time reductions for kinematic modeling and motion planning using the Cosserat rod model \cite{bentley2022interactive}.

Rao, et al., perform a comparison of the different physics-based models and provide a set of guidelines for determining which model to use based on the structure of the robot assuming parallel routed tendons \cite{rao2021model}. They conclude that when the structure of the robot contains a small number of separator disks holding the tendons, PRBM provides the best accuracy with respect to its computation time; however, as the number of disks increases, it becomes more useful to use VC models. We compare our work in this paper to the state-of-the-art Cosserat rod model described in \cite{bentley2022interactive}.

\begin{figure*}
    \centering
    \begin{subfigure}{\columnwidth}
        \centering
        \includegraphics[width=\linewidth]{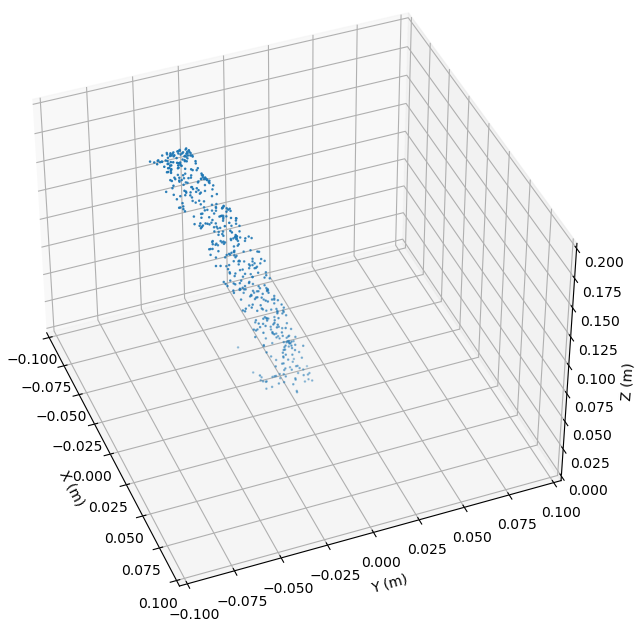}
        \label{subfig:SinglePCD}
    \end{subfigure}
    \begin{subfigure}{\columnwidth}
        \centering
        \includegraphics[width=\linewidth]{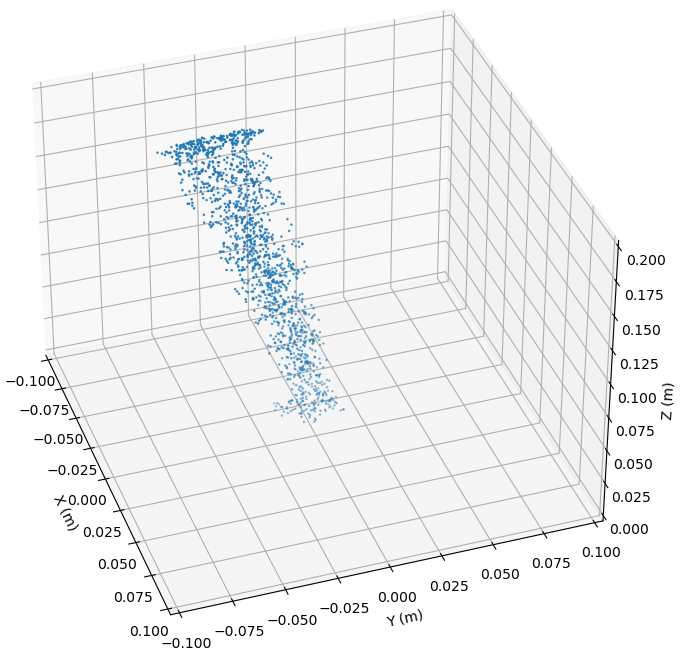}
        \label{subfig:FullPCD}
    \end{subfigure}
    \caption{An example configuration in the training data set. (Left) One particular instance of the configuration. (Right) The concatenated distribution of all point clouds gathered during data collection illustrating the inherent uncertainty associated with the geometry of the robot at a given configuration.}
    \label{fig:DataSetExample}
\end{figure*}

\subsection{Data-Driven Models}
Data-driven models have also been proposed to model both the kinematics and dynamics for continuum manipulators. Nonlinear auto-regressive models with exogeneous inputs have been shown to perform well in predicting the end effector position for continuum robots \cite{parvaresh2022dynamics}. While this model is able to implicitly learn how to model the forces on the robot, it does not reason over the uncertainties of continuum robots' geometries. 

Reinhart and Steil compare several data-driven and hybrid modeling approaches for the forward-kinematics of soft robots \cite{reinhart2016hybrid}. They concluded that a non-linear Extreme Learning Machine (ELM) achieved the best result for minimizing the error with respect to the end effector position.

Work has also been done to learn the full geometry for concentric tube robots as a parametric curve \cite{kuntz2020learning}. Their model is capable of achieving high accuracy for concentric tube robots but does not allow for reasoning over the geometric uncertainties associated with continuum robots.

Others propose to use Gaussian Process Regression (GPR) for the forward kinematics of tendon-driven continuum robots \cite{shen2020accuracy}. They show a low position and orientation error post-training; however, this method does not allow for reconstruction of the full robot geometry. Gaussian Mixture Regression (GMR) has also been used for the inverse kinematics \cite{chen2016learning}. They find that Gaussian mixtures do not perform as well as other methods for computing the inverse kinematics of tendon-driven continuum robots. By contrast, we find in this paper that Gaussian mixtures perform well with respect to forward kinematics for representing the full geometry of the robot.

The current state of data-driven approaches to forward kinematic models primarily focuses on learning how to predict the position of the end effector of the robot while implicitly modeling the forces acting on the robot. The approach proposed in this paper extends these ideas to modeling the entire geometry of the robot as a Gaussian mixture model. Unlike prior work, this new model is capable of modeling the geometric uncertainties of tendon-driven continuum robots. In our experiments, we show that we are capable of explicitly reasoning over these geometric uncertainties to improve the safety of tendon-driven continuum robot operation.

\section{Kinematic Modeling}
We propose a novel learned, data-driven approach to modeling the kinematics for tendon-driven continuum robots. As opposed to standard kinematic models that directly compute the position and orientation of each joint axis of the robot, our model computes a Gaussian mixture model (GMM) representation of the robot's geometry in the workspace. A GMM is a weighted combination of Gaussian probability distributions. The probability density function of a GMM for a random variable $x$ is formulated as
\begin{equation}
\begin{gathered}
p(x)=\sum_{i=1}^{n} w_ip_i(x) \\
p_i(x)=\frac{\exp{(-\frac{1}{2}(x-\boldsymbol{\mu}_i)^T\Sigma_{i}^{-1}(x-\boldsymbol{\mu}_i))}}{\sqrt{(2\pi)^n|\Sigma_i|}} \\
\end{gathered}
\end{equation}
for $n$ Gaussian components
where $w_i\geq0$ and $\sum_{i=1}^n w_i = 1$,
and $\boldsymbol{\mu}_i$, $\Sigma_i$, and $w_i$ are the mean, covariance matrix, and weight for component $i$ respectively.

Given $d=(d_1, d_2, ..., d_m)$, where $d_j$ is the displacement of tendon $j$, the model computes the GMM through a mixture density network (MDN). MDNs are a type of neural network composed of an initial feedforward fully connected network followed by separate feedforward fully connected networks to compute the means, covariance matrices, and weights for each component of the GMM. Figure \ref{fig:MDNArchitecture} shows the architecture used to model the kinematics for the tendon-driven continuum surgical manipulator. For the purposes of modeling the workspace geometry of the robot, the MDN produces a three dimensional GMM for each configuration of tendon displacements.

We note that the model computes the matrices $\mathrm{U}_i$ as opposed to the covariance matrices $\Sigma_i$. Following the derivation from \cite{kruse2020technical}, $\mathrm{U}_i$ is used to reconstruct the precision matrix (inverse covariance matrix) for component $i$.
\begin{equation}
(\mathrm{\Bar{U}}_i)_{jk} = \begin{cases}
    (\mathrm{U}_i)_{jk}, & $if $ j \neq k \\
    \exp(\mathrm{U}_i)_{jk}, & $otherwise$
\end{cases}
\end{equation}
where $\mathrm{\Bar{U}}_i$ is the upper triangular matrix of the Cholesky decomposition of the precision matrix for component $i$, and indices $j,k$ are the row and column for the corresponding matrix respectively. Thus, we have
\begin{equation}
    \Sigma_i^{-1} = \mathrm{\Bar{U}}_i^T\mathrm{\Bar{U}}_i
    \label{eq:precision}
\end{equation}

This enables the outputs of the MDN to be unconstrained while maintaining that the covariance matrix for each component is positive-definite. The negative log-likelihood function can also be efficiently computed using $\mathrm{U}_i$ which reduces the required training time. This results from the ability to compute the log of the square root of the determinant of the precision matrix as the sum of the diagonal elements of $\mathrm{U}_i$ \cite{kruse2020technical}. Following from equation \ref{eq:precision},

\begin{equation}
|\Sigma_i^{-1}| = \biggl(\prod_{j=1}^3\diag(\mathrm{\Bar{U}}_i)_j\biggr)^2.
\label{eq:prec_det}
\end{equation}
Taking the square root gives
\begin{equation}
|\Sigma_i^{-1}|^{\frac{1}{2}} = \prod_{j=1}^3\diag(\mathrm{\Bar{U}}_i)_j.
\end{equation}
We can then take the log to result in
\begin{equation}
\log{|\Sigma_i^{-1}|^{\frac{1}{2}}} = \sum_{j=1}^3\diag(\mathrm{U}_i)_j.
\end{equation}

The MDN is trained to minimize the negative log-likelihood of ground-truth data collected from the robot. To gather the ground-truth distributions, we collect point clouds of the robot's geometry at various configurations using two RealSense cameras on opposing sides of the robot. We chose configurations for the data set by discretizing the robot's configuration space and collecting data for all configurations in this discretization. For each configuration in the data set, we randomly sample multiple starting configurations of the robot, then move the robot along a linear interpolation in the configuration space between the starting and desired configuration. A point cloud is then collected once the robot has arrived at the desired configuration. The concatenated point clouds across the different starting configurations gives a distribution of potential geometries the robot can have at the desired configuration. Figure \ref{fig:DataSetExample} shows an example of one configuration in the training set. The right graph of figure \ref{fig:DataSetExample} illustrates the variability that arises in the robot's resulting geometry despite being in the same configuration. The training and testing data sets contain 2,277, and 253 configurations respectively.

Given the training set of point clouds X and corresponding tendon displacement configurations Y, we produce the associated GMM for each configuration $y\in Y$ using the MDN.

\begin{figure}
    \centering
    \includegraphics[width=\columnwidth]{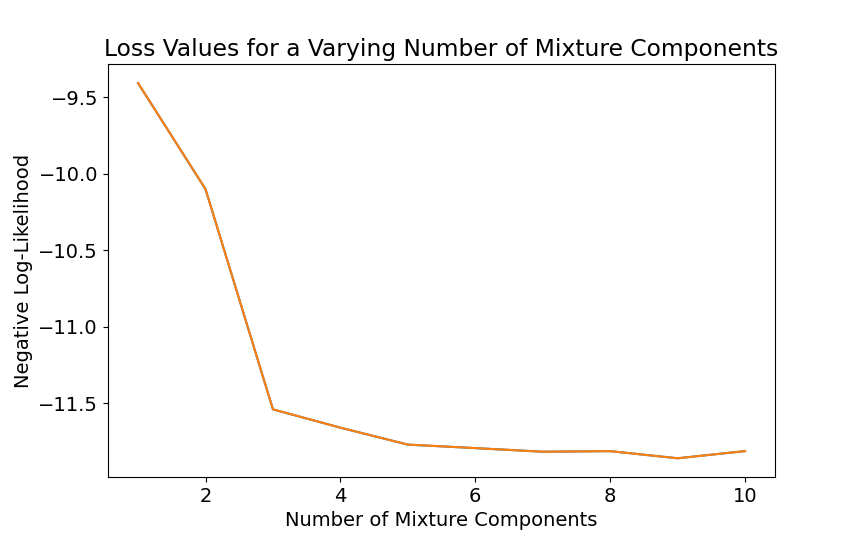}
    \caption{Loss function values at different numbers of mixture components on the test data set. The results suggest that 5 mixture components are the minimum number of components necessary to sufficiently minimize the negative log-likelihood.}
    \label{fig:likelihood}
\end{figure}

\begin{equation}
\centering
\begin{aligned}
    MDN_\theta(y) = w_{i|\theta}, \boldsymbol{\mu}_{i|\theta}, \mathrm{U}_{i|\theta} \\
     i \in \{1, 2, ..., n\}
\end{aligned}
\end{equation}
where $\theta$ are the network weights of the MDN.

Using the computed GMMs, the negative log-likelihood of the training set point clouds is computed using the loss function derived in \cite{kruse2020technical}.
\begin{equation*}
\centering
\begin{aligned}
    \loss(\theta) &= \EX_{x \in X, y \in Y}[-\log{p_\theta(x|y)}] \\
    &\propto \EX_{x \in X, y \in Y}\biggl[-\log\sum_{i=1}^n w_{i|\theta} \cdot \exp\biggl(\sum_{j=1}^3\diag(\mathrm{U}_{i|\theta})_j \\ &-\tfrac{1}{2} {\|\mathrm{\Bar{U}}_{i|\theta}(x-\boldsymbol{\mu}_{i|\theta})\|}_2^2\biggr)\biggr]
\end{aligned}
\end{equation*}
The network parameters $\theta$ are optimized during training to minimize $\loss(\theta)$. Figure \ref{fig:MainFig} shows the training process for the MDN.

\begin{figure}
    \centering
    \begin{subfigure}{\columnwidth}
        \centering
        \includegraphics[width=0.7\columnwidth]{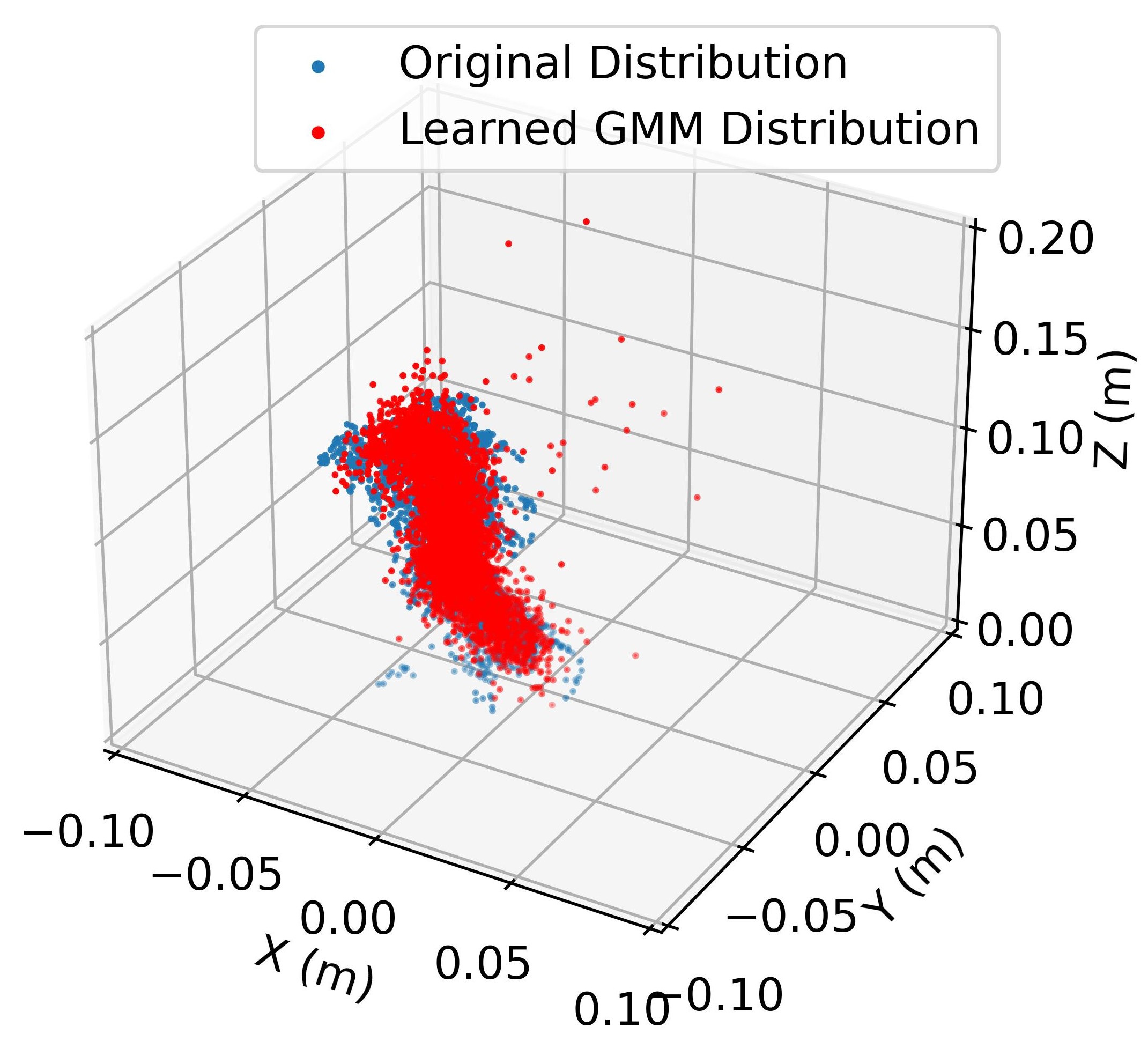}
    \end{subfigure}
    \begin{subfigure}{\columnwidth}
        \centering
        \includegraphics[width=0.7\columnwidth]{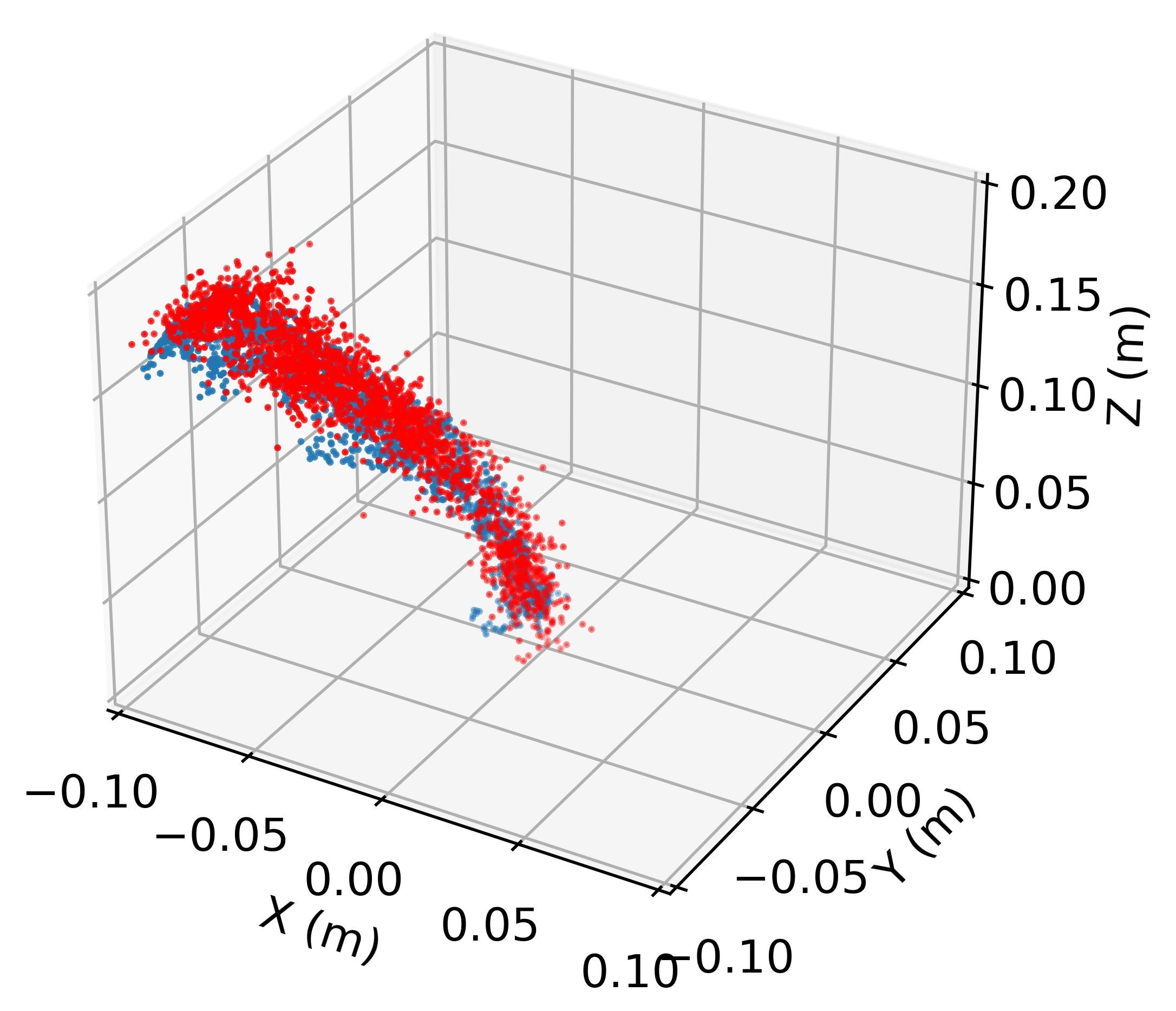}
    \end{subfigure}
    \caption{An example set of outputs from the mixture density network after training with five mixture components compared with the 
    ground-truth
    geometric distribution. The red points are sampled from the Gaussian mixture model output from the mixture density network. The blue points are the points collected from the robot during data collection. The sampled points demonstrate how the learned Gaussian mixture model is capable of representing the full distribution of geometries as a result of the kinematic uncertainty.}
    \label{fig:GMMResults}
\end{figure}

\section{Analysis}
We trained our model on a tendon-driven continuum robot with nine separator disks. The robot is 0.2 m in length and is controlled via three parallel routed tendons and one helical tendon. 

The model was trained with the number of mixture components (Gaussians) ranging from one to ten. Figure \ref{fig:likelihood} shows the negative log-likelihood values on the test data set on models with different numbers of mixture components.
We find that five mixture components minimize the negative log-likelihood the most while avoiding mode collapse. Mode collapse occurs when the MDN begins to output one or multiple component mixture weights arbitrarily close to zero.
This implies that the model could have been trained with fewer mixture components while still achieving the same loss. Having a number of mixture components greater than five leads to this mode collapse phenomenon.

We note that the optimal number of mixture components is most likely robot specific.
For different structures of the robot with different tendon routing patterns, it will be necessary to experimentally determine the appropriate number of mixture components for the GMM.
We leverage the model with five components for the rest of the evaluation.

\begin{figure}
    \centering
    \begin{subfigure}{\columnwidth}
        \centering
        \includegraphics[height=200px]{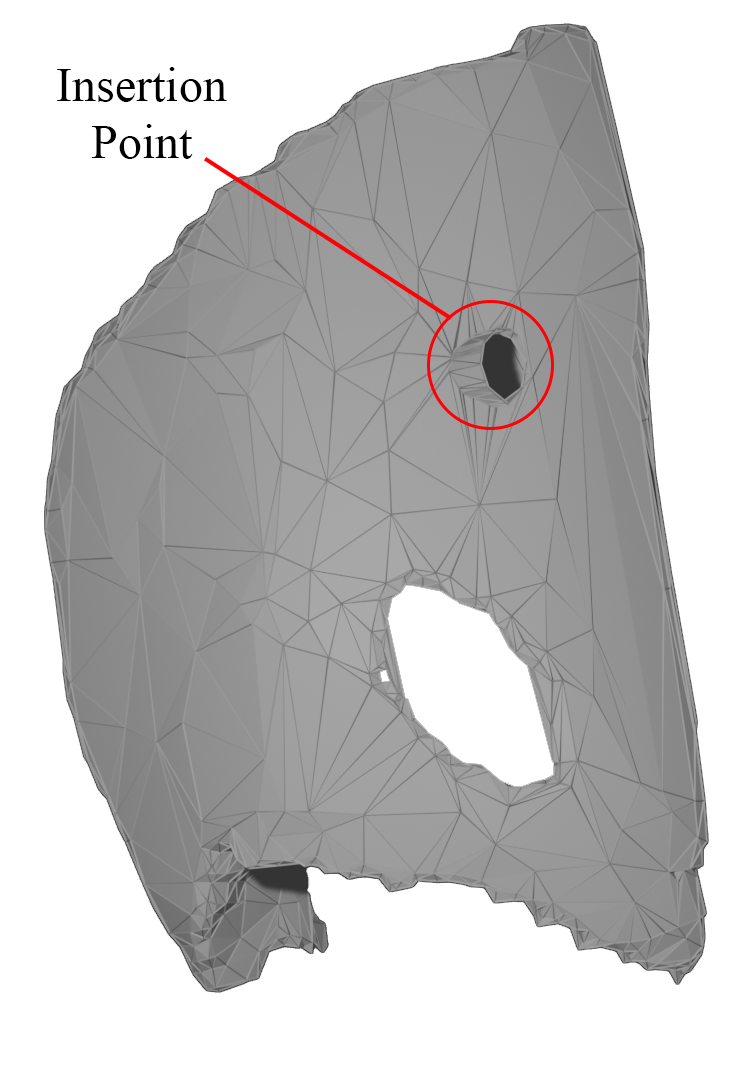}
    \end{subfigure}
    \begin{subfigure}{\columnwidth}
        \centering
        \includegraphics[width=\columnwidth]{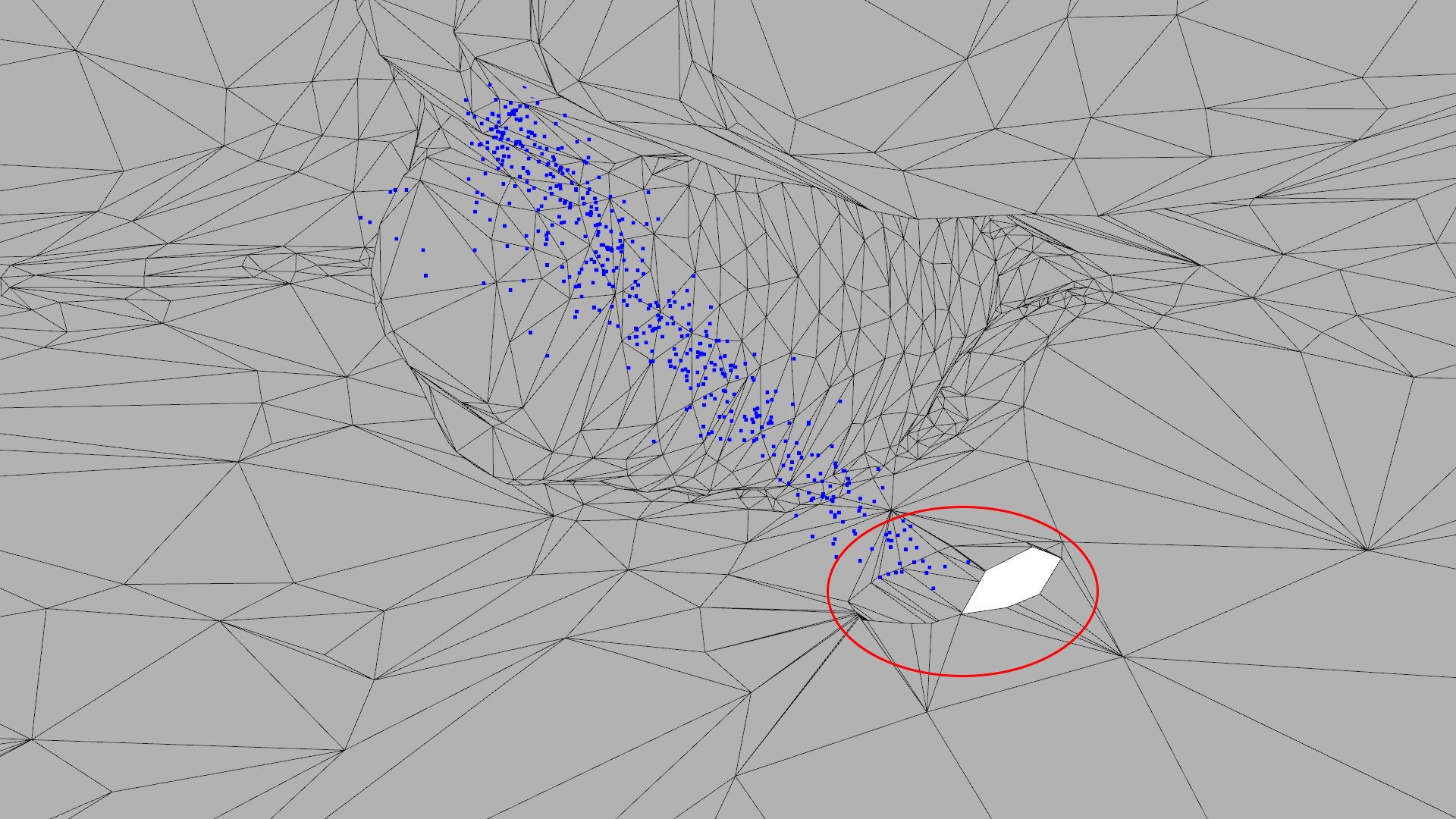}
    \end{subfigure}
    \caption{The simulated chest cavity mesh environment used during motion planning, segmented from a patient undergoing a procedure to treat pleural effusion. (Top) The exterior of the volume. The red circle shows the insertion point for the robot. (Bottom) The interior of the volume. The red circle shows the insertion point for the robot. The blue points show a point cloud representation of the robot's shape inside the chest cavity in its home position (zero displacement for all tendons).}
    \vspace{-1.5em}
    \label{fig:LungEnv}
\end{figure}

Figure \ref{fig:GMMResults} shows the model's output at two different randomly sampled configurations of the robot compared with the ground-truth geometric distribution at those configurations. The points sampled from the GMM accurately represent the expected geometry of the robot with a high density of points being sampled within the ground-truth distribution and a low density of points being sampled everywhere else.

We compare the computation time of the trained MDN with the Cosserat rod model presented in \cite{bentley2022interactive}. Timing our model over 10,000 randomly sampled configurations results in a mean computation time of 0.33 ms. The results in \cite{bentley2022interactive} show a mean computation time of 0.39 ms for the Cosserat rod model. This demonstrates a 15\% reduction in computation time from the Cosserat rod model to our model.

We note that while standard comparisons of different kinematic models for tendon-driven continuum robots directly compare the accuracy of each model with respect to the predicted endpoint, we cannot provide this comparison for our model. Because our model learns distributions over geometries, there is no inherent notion of a predicted endpoint which poses a fundamental difference between the proposed model and standard kinematic models.

\section{Planning with Gaussian Kinematics}
One application for this kinematic model is safe motion planning.
Using the learned kinematic model enables explicit reasoning over the probability of collision with obstacles as a result of the geometric uncertainty of the robot.
Other motion planners that consider the probability of collision primarily consider the uncertainty as a result of motion and sensing uncertainty \cite{van2011lqg,van2012motion,sun2016safe}.
Our model enables an extension of these motion planners to reason explicitly over the geometric uncertainty associated with the kinematics of tendon-driven continuum surgical manipulators. Here we present a motion planner that uses our learned kinematic model to explicitly minimize the probability of collision for a given nominal trajectory of tendon displacements.

We first construct a nominal trajectory using the Probabilistic Roadmap (PRM)~\cite{kavraki1996probabilistic} motion planning algorithm, a widely used algorithm in robotics and motion planning to find feasible paths for robots operating in complex environments.
It does so by constructing a roadmap of the given environment via random sampling in the robot's configuration space, enabling the robot to efficiently navigate to a goal while avoiding obstacles.
However, any planning algorithm could be used to generate a nominal trajectory of configurations.
In the generation of the initial roadmap (and subsequently the nominal trajectory), we check for collisions by testing that the means of each component of the GMM are not in collision with the obstacles in the environment.
This condition enables us to later efficiently compute the probability of collision with the environment.

To improve the safety of the motion plan, we locally optimize the nominal trajectory itself with respect to the probability of collision as determined by our model.
The probability of collision along the whole trajectory is minimized using Bayesian Optimization.
To efficiently compute the probability of collision, we extend the method in \cite{patil2012estimating} to reason over the probability of collision between a GMM and a given environment.
To do so, for each component in the predicted GMM, we transform the environment's geometry such that the mean of the mixture component is at the origin with a covariance matrix equal to the identity matrix. We translate the environment by the negative of the mean and multiply by the Cholesky decomposition of the precision matrix. The component distribution in the new environment is the unit sphere centered at the origin. The new environment is constructed via

\begin{equation}
\centering
    \mathrm{\hat{E}}_i = \mathrm{\Bar{U}}_i (\mathrm{E} - \boldsymbol{\mu}_i)
\end{equation}
where $\mathrm{E}$ is the set of all vertices in a mesh representation of the environment and $\mathrm{\hat{E}}_i$ is the set of all mesh vertices in the transformed environment for mixture component $i$.

A locally convex region of free space is then defined over this transformed environment using the method presented in \cite{patil2012estimating} as a set of linear constraints $\boldsymbol{a}_{ij}^T \boldsymbol{p} \leq b_{ij}$, where $\boldsymbol{a}_{ij}, \boldsymbol{p} \in \mathbb{R}^3$, $b_{ij} \in \mathbb{R}$, $i$ is the mixture component index, $j$ is the constraint index, and $\boldsymbol{p}$ is a point in the robot's workspace. A conservative estimate for the probability that a configuration $\boldsymbol{x}$ is not in free space $\mathrm{X}_f$ is computed by
\begin{equation}
    p(\boldsymbol{x} \not\in \mathrm{X}_f) \leq \sum_{i=1}^N\biggl(w_i\sum_{j=1}^{K_i} (1-\texttt{cdf}(b_{ij}))\biggr)
\end{equation}
for $N$ mixture components and where $K_i$ is the number of linear constraints for component $i$. The \texttt{cdf} function here refers to the cumulative distribution function for the standard normal Gaussian $\mathcal{N}[0,1]$. The probability of collision along a trajectory $\tau$ is estimated by taking the complement of the product of the probability that each state in the trajectory is not in collision.
\begin{equation}
    p(\exists \boldsymbol{x} \in \tau:x\not\in \mathrm{X}_f) \leq 1 - \prod_{\boldsymbol{x}\in\tau} (1 - p(\boldsymbol{x} \not\in \mathrm{X}_f))
    \label{eq:obj}
\end{equation}

Using Bayesian Optimization~\cite{pelikan1999boa}, the upper bound on $p(\exists \boldsymbol{x} \in \tau:\boldsymbol{x}\not\in \mathrm{X}_f)$ is iteratively minimized for the given nominal trajectory. During each iteration of the optimization, we choose one intermediate configuration along the trajectory to optimize.
New potential configurations are randomly sampled in a neighborhood about the chosen configuration.
We evaluate each random sample using the probability of improvement acquisition function~\cite{jones2001taxonomy}.
The probability of collision is then recomputed using the sample with the highest likelihood of reducing the probability of collision.
If the sampled trajectory has a lower probability of collision than the previous best trajectory, the best trajectory is updated to the sample.
This process is repeated for a predefined number of iterations with a predefined number of samples at each iteration.

\begin{figure}
    \centering
    \includegraphics[width=\columnwidth]{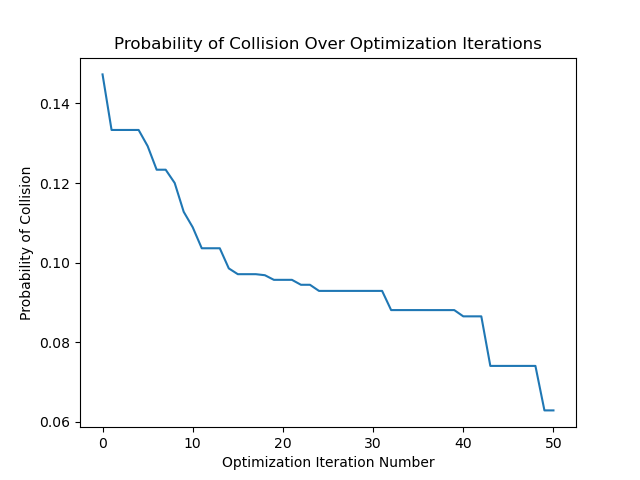}
    \caption{Results from the optimization of a nominal trajectory in the simulated chest cavity mesh environment. We ran 50 iterations of Bayesian Optimization with respect to the probability of collision. The nominal trajectory had approximately a 15\% chance of collision prior to optimization. After optimization, the probability of collision had been reduced to under 7\%.}
    \vspace{-1.5em}
    \label{fig:OptimResults}
\end{figure}

We tested the motion planner in a simulated chest cavity mesh environment shown in figure \ref{fig:LungEnv}. The nominal trajectory through the chest had a 15\% chance of colliding after planning using the PRM. After 50 iterations of optimization, the probability of collision had been reduced to under 7\%. Figure \ref{fig:OptimResults} shows the change in probability of collision over the optimization iterations. Using our learned kinematic model, we more than halved the probability of collision.

\section{Conclusion}
We present a novel approach to the modeling of tendon-driven continuum robot kinematics. Our method explicitly models the geometric uncertainties associated with the robot kinematics through a mixture density network trained to output Gaussian mixture models that accurately reflect the distribution over geometries. We find that our model accurately represents the ground-truth distributions over geometries and is computationally faster than a current state-of-the-art Cosserat rod model. We also demonstrate the capabilities of our model in the downstream task of motion planning. We plan nominal trajectories and use the trained mixture density network to explicitly minimize the probability of collision using Bayesian Optimization.

In the future we intend to apply the model to other continuum robot types.
We also intend to integrate the optimization into the sampling process of a motion planner that integrates sampling with optimization for motion planning beyond the optimization of nominal trajectories, similar to as in~\cite{Kuntz2020_ISRR,Kuntz2019_IROS} but with our uncertainty model and metric.

We believe that the GMM-based kinematic model presented in this work has the potential to improve safe planning and control of tendon-driven continuum robots in medicine.

\bibliography{Citations}
\bibliographystyle{IEEEtran}

\end{document}